\documentclass[10pt,twocolumn,letterpaper]{article}

\usepackage{cvpr}              

\usepackage[dvipsnames]{xcolor}

\newcommand{\cmark}{\ding{51}}%
\newcommand{\xmark}{\ding{55}}

\usepackage{amsmath,amsfonts,bm}

\def\eqref#1{equation~\ref{#1}}

\def\1{\bm{1}}

\def\vq{{\bm{q}}}

\def\mK{{\bm{K}}}

\def\mQ{{\bm{Q}}}

\def\mS{{\bm{S}}}

\def\mV{{\bm{V}}}

\DeclareMathAlphabet{\mathsfit}{\encodingdefault}{\sfdefault}{m}{sl}
\SetMathAlphabet{\mathsfit}{bold}{\encodingdefault}{\sfdefault}{bx}{n}

\newcommand{\vpar}[1]{\vspace{3mm}\noindent\textbf{#1}\ \ }

\newcommand{\ignorethis}[1]{}

\makeatletter
\DeclareRobustCommand\onedot{\futurelet\@let@token\@onedot}
\def\@onedot{\ifx\@let@token.\else.\null\fi\xspace}

\makeatother

\definecolor{mydarkblue}{rgb}{0,0.08,1}
\definecolor{mydarkgreen}{rgb}{0.02,0.6,0.02}
\definecolor{mydarkred}{rgb}{0.8,0.02,0.02}
\definecolor{mydarkorange}{rgb}{0.40,0.2,0.02}
\definecolor{mypurple}{RGB}{111,0,255}
\definecolor{myred}{rgb}{1.0,0.0,0.0}
\definecolor{mygold}{rgb}{0.75,0.6,0.12}
\definecolor{myblue}{rgb}{0,0.2,0.8}
\definecolor{mydarkgray}{rgb}{0.66,0.66,0.66}

\definecolor{freecolor}{rgb}{0.96,0.68,0.16}
\definecolor{frozoncolor}{rgb}{0.39,0.41,0.41}

\def\loss{\mathcal{L}\xspace}

\def\weight{\mathbf{w}\xspace}
\def\act{\mathbf{a}\xspace}
\def\inputgrad{\frac{\partial \loss}{\partial \act_i}\xspace}

\def\weightgrad{\frac{\partial \loss}{\partial \weight_i}\xspace}

\definecolor{cvprblue}{rgb}{0.21,0.49,0.74}
\usepackage[pagebackref,breaklinks,colorlinks,citecolor=cvprblue]{hyperref}

\def\paperID{44} 
\def\confName{CVPR}
\def\confYear{2024}

\title{Block Selective Reprogramming for On-device Training of Vision Transformers}
\author{Sreetama Sarkar$^{1}$ \ \ \ \
Souvik Kundu$^{2}$ \ \ \ \ 
Kai Zheng$^{1}$ \ \ \ \
Peter A. Beerel$^{1}$\\
$^{1}$Universiy of Southern California, Los Angeles, USA \ \ \ \ 
$^{2}$Intel Labs, USA \\
{\tt\small {\{sreetama,kzheng44,pabeerel\}@usc.edu} \ \ \ \ \tt\small {souvikk.kundu}@intel.com}
}
\begin{document}
\maketitle
\begin{abstract}
 The ubiquity of vision transformers (ViTs) for various edge applications, including personalized learning, has created the demand for on-device fine-tuning. However, training with the limited memory and computation power of edge devices remains a significant challenge. 
In particular, the memory required for training is much higher than that needed for inference, primarily due to the need to store activations across all layers in order to compute the gradients needed for weight updates.
Previous works have explored reducing this memory requirement via frozen-weight training as well storing the activations in a compressed format.
However, these methods are deemed inefficient due to their inability to provide training or inference speedup. 
In this paper, we first investigate the limitations of existing on-device training methods aimed at reducing memory and compute requirements.
We then present \textit{block selective reprogramming} (BSR) in which we fine-tune only a fraction of total blocks of a pre-trained model 
and selectively drop tokens based on self-attention scores of the frozen layers. 
To show the efficacy of BSR, we present extensive evaluations on ViT-B and DeiT-S with five different datasets. Compared to the existing alternatives, our approach simultaneously reduces training memory by up to $1.4\times$ and compute cost by up to $2\times$ while maintaining similar accuracy. We also showcase results for Mixture-of-Expert (MoE) models, demonstrating the effectiveness of our approach in multitask learning scenarios.
\end{abstract} 

\section{Introduction}
Over the past several years, there has been an unprecedented growth in the deployment of deep learning applications on edge devices. Mutitask learning (MTL) is gaining momentum in edge applications due to their ability to dynamically adapt to different tasks with minimum overhead. Currently, edge devices primarily handle inference, while the training or fine-tuning processes take place in the cloud. This not only involves substantial communication overhead for transferring both the model and data, but also raises concerns about data privacy \cite{kundu2023learning}.  Although on-device training is preferred, memory limitations of edge devices pose challenges. Naive solutions include directly using pre-trained models or fine-tuning only the last layer. However, these approaches can lead to a notable drop in accuracy if the distribution of the new data differs significantly from the pre-training data. 

\begin{figure}[!t]
    \centering
    \includegraphics[width=0.45\textwidth]{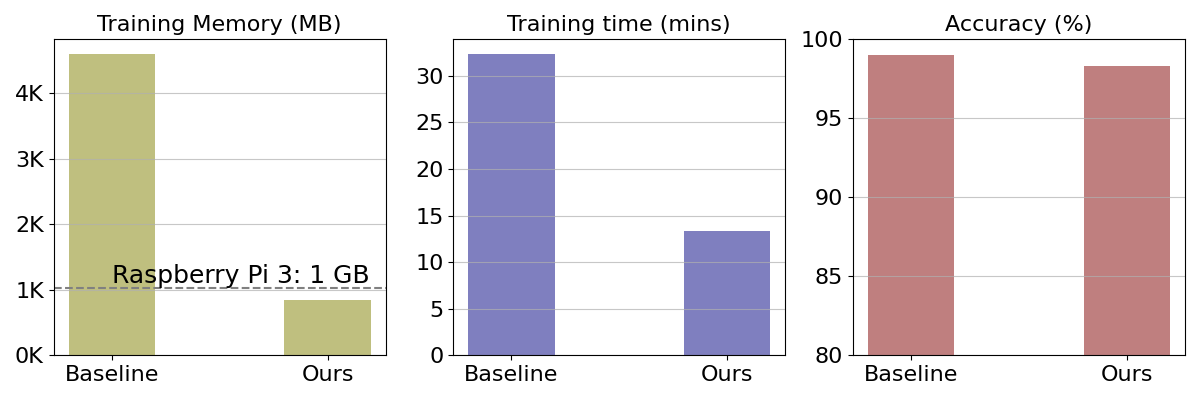}
    \caption{Test accuracy, training time, and memory comparison for a ViT-B on CIFAR-10 with a batch size of 32. In particular, we achieve a 5.47$\times$ memory reduction and 2.43$\times$ training speedup while yielding similar test accuracy. Our benefits are even more significant for higher batch sizes.}
    \label{fig:intro_fig}
\end{figure}
Previous works on efficient on-device learning \cite{cai2020tinytl, jiang2022back} have pointed out that the training memory is primarily dominated by activations, rather than parameters. In order to reduce activation memory, training residual modules while keeping the original backbone network frozen, has been widely explored in convolutional neural networks (CNNs) \cite{cai2020tinytl, yang2022rep}. However, this approach is not effective in vision transformers (ViTs) due to the presence of non-linear layers such as self-attention, softmax, and GELU, that require storing the input activation for gradient computation, even with frozen weights. Another line of research has explored reducing the memory cost of ViTs by introducing irregular sparsity in activation tensor \cite{jiang2022back}. However, such approaches might not yield significant memory savings due to the overhead of sparse representations and do not provide any potential speedup.

In contrast, this work presents a memory and parameter-efficient fine-tuning approach called block selective reprogramming (BSR) that fine-tunes only a fraction of the pre-trained weight tensors such that the activation memory requirement to propagate the gradients is minimized. 

\noindent\textbf{Our Contributions}
We first investigate and identify the limitations of existing SoTA methods applied to CNNs in the context of ViTs, as well as the shortcomings of existing on-device ViT fine-tuning approaches. In particular, we observe that residual fine-tuning does not necessarily reduce activation memory and can suffer from increased computational overhead. 

Based on our observation, we propose BSR where we selectively fine-tune a small fraction of blocks of a pre-trained network. 
Additionally, to reduce the compute cost, we employ a token-dropping method that drops tokens primarily based on the self-attention score of the frozen layers. This saves activation memory and yields training speedup, unlike activation sparsification \cite{jiang2022back}.
 
We present a detailed empirical evaluation of the impact of the block selection choice and token drop locations on accuracy. We demonstrate the efficacy of BSR on two different transformer models with five different datasets. Specifically, compared to the existing alternatives, our approach yields additional training memory
saving of up to $1.4\times$ while saving compute cost by up to $2\times$. Compared to fully fine-tuning an ImageNet pre-trained ViT-B \cite{dosovitskiy2020image} model on CIFAR-10 (Figure \ref{fig:intro_fig}), our approach provides $5.47 \times$ training memory reduction along with $2.43\times$ reduction in training time for a batch size of 32.

Finally, we consider multi-task learning (MTL) as a target for our fine-tuning approach because of the benefits of maintaining a single model across different tasks in edge applications. In particular, mixture-of-expert (MoE) models \cite{chen2023sparse, liang2022mvit} are becoming popular in MTL due to their ability to disentangle the parameter space yielding improved MTL performance. We achieve close to baseline accuracy for fine-tuning MoE using BSR while reducing training memory and FLOPs by 2.3$\times$ and 1.5$\times$ respectively, demonstrating the effectiveness of our approach in MTL scenarios.

\section{Related Work}
\noindent\textbf{On-device Training} TinyTL \cite{cai2020tinytl} was one of the first papers to show that storing activation, not parameters, is the primary bottleneck for on-device training. Therefore, model compression techniques like weight pruning \cite{han2015learning, he2017channel, frankle2018the} and quantization \cite{han2016deep, courbariaux2015binaryconnect} do not significantly reduce the cost of on-device training. Parameter efficient training methods, like Bitfit \cite{bitfit} or training only Batch Normalization parameters \cite{ioffe2015batch}, reduces parameter memory, which is only a small fraction of the total train memory \cite{cai2020tinytl, jiang2022back}. Recomputing discarded activations during back-propagation \cite{gruslys2016memory, chen2016training} reduces memory cost at the expense of large computational overhead, making them unsuitable for edge devices. TinyTL proposes training only biases with frozen weights. To compensate for reduced learning capacity, they introduce residual blocks, which are much smaller in width and employ group convolution to reduce memory footprints. Rep-Net \cite{yang2022rep} improves on this residual learning approach by designing six small residual modules which exchange features intermittently with a frozen pre-trained network, achieving a lower memory footprint and improved accuracy. However, introducing residual blocks leads to an increase in parameters, as well as both training and inference compute. Other methods for activation memory reduction include activation quantization \cite{evans2021ac, pan2021mesa, kundu2022bmpq} and pruning \cite{NEURIPS2019_30c8e1ca, jiang2022back, kundu2021dnr, kundu2022toward}. Back-Razor \cite{jiang2022back} proposes pruning the activation stored for backpropagation after the forward pass. They sparsify activations up to 95\%, yielding a $5.5\times$ reduction in training memory for ViTs. Their training memory reduction is limited by the memory required to save a binary mask of the sparse activations and hence, cannot achieve memory reduction proportional to activation sparsity. Notably, none of the current on-device learning methods provide speedup or computational efficiency during inference.   

\vpar{Efficient Vision Transformers} 
Design for efficient ViTs is an active area of research and can be partitioned into two broad categories, namely architecture design for efficient ViTs and optimization of the operations for ViTs. In the first category, researchers have explored different self-attention alternatives including Linformer \cite{wang2020linformer}, SAL-ViT \cite{zhang2023sal}, and Performer \cite{choromanski2020rethinking}, that targeted reducing its quadratic compute complexity. Other works presented resource and latency friendly efficient architectures including Mobile-former \cite{chen2022mobile} and Efficientformer \cite{li2022efficientformer}. However, these works do not target on-device fine-tuning.
For the second category, researchers have developed various compression \cite{yu2022unified, kundu2023sensi, yin2023junk}, token dropping \cite{liang2022not}, and token merging \cite{bolya2022token} schemes to yield compute efficiency which are designed to provide inference speedup. However, despite reducing compute and often latency, such methods alone cannot yield the desirable reduction in activation and gradient storage necessary for on-device fine-tuning. In this paper, we leverage token dropping for on-device fine-tuning with resource-constrained activation memory and compute budget.



\section{Motivational Analysis}
\subsection{Preliminaries}
ViT models partition the input image into $N$ patches, also called tokens, and embed each token into an embedding vector of length $L$. An extra classification token $cls\_token$ is added to the set of image tokens, creating $N+1$ tokens. The $cls\_token$ is responsible for aggregating global image information. A trainable positional embedding is added to each of the token embeddings, which are then passed through a series of transformer encoder blocks, each consisting of a multi-head self-attention (MHSA) layer followed by a feed forward network (FFN). The tokens are mapped into Query ($\mQ$), Key ($\mK$), and Value ($\mV$) matrices, each having a dimension of $\left[N+1, d\right]$, where $d=L/H$ and $H$ is the number of heads in the MHSA. Each head performs the self-attention operation 
\begin{equation}
    {\rm Attention}(\mQ, \mK, \mV) = {\rm Softmax}(\frac{\mQ \mK^T}{\sqrt{d}}) \mV.
    \label{eq:attn}
\end{equation}
The FFN is usually a 2-layer network with GELU activation and a hidden dimension denoted $D_{ffn}$.

\subsection{Analyzing Memory Cost}
We perform a detailed analysis of the activation memory required for gradient computation in a ViT block. The gradient for back-propagation for linear and non-linear layers are given by Equation \ref{eq:linear_backward} and Equation \ref{eq:non_linear_backward} respectively, as follows. 
\begin{align}
    \label{eq:linear_backward}
    \inputgrad = \frac{\partial \loss}{\partial \act_{i+1}} \frac{\partial \act_{i+1}}{\partial \act_i} =& \frac{\partial \loss}{\partial \act_{i+1}} \weight_i^T \text{\hspace{2pt}for linear layers}\\
    \label{eq:non_linear_backward}
     =& \frac{\partial \loss}{\partial \act_{i+1}} \mathbf{h}\left(\act_{i}\right) \text{\hspace{2pt}for non-linear layers}
\end{align}
Here, the input and output activations for the $i^{th}$ layer are denoted by $\act_{i}$ and $\act_{i+1}$. For linear layers, $\act_{i+1} =  \weight_i\act_{i}+b$ and the gradient $\frac{\partial \act_{i+1}}{\partial \act_i}$ is independent of $\act_i$, whereas, for non-linear layers, the gradient is a function of $\act_i$, given by $\mathbf{h}(\act_i)$. The gradient for the weights are given by Equations \ref{eq:linear_wt_grad} and \ref{eq:non_linear_wt_grad} where $\mathbf{g}$ denotes the gradient function with respect to weights for non-linear layers, as follows.
\begin{align}
    \label{eq:linear_wt_grad}
    \weightgrad = \frac{\partial \loss}{\partial \act_{i+1}} \frac{\partial \act_{i+1}}{\partial \weight_i} =& \frac{\partial \loss}{\partial \act_{i+1}} \act_i^T \text{\hspace{2pt}for linear layers}\\
    \label{eq:non_linear_wt_grad}
     =& \frac{\partial \loss}{\partial \act_{i+1}} \mathbf{g}\left(\act_{i}\right) \text{\hspace{2pt}for non-linear layers}
\end{align}
The gradients of weights are  dependent on input activations, whether the layer is linear or non-linear. Thus, trainable blocks must store activations both for linear layers, including LayerNorm or fully-connected layers, and non-linear layers, including Softmax, GELU, and Attention. In contrast, frozen blocks situated in the gradient flow path only need to store activations for non-linear layers. The memory cost for the stored activations for non-linear layers in DeiT-S \cite{deit} are presented in Table \ref{tab:memory_cost}. A detailed analysis of the gradient activation memory required by each block reveals that a frozen block stores ${\sim}2.9$ MB whereas a fully trainable block stores ${\sim}5.5$ MB, indicating that training with frozen weights reduces memory requirements by ${\sim2}\times$.
\begin{table}[h]
\small\addtolength{\tabcolsep}{-3pt}
    \begin{center}
        \begin{tabular}{cccc}
        \toprule
 		\textbf{Module} & \textbf{Stored} & \textbf{Dimension} & \textbf{Memory} \\
   & \textbf{Activation} & & \textbf{Cost (MB)} \\

 		\midrule
   Self-Attention & $\mQ, \mK, \mV$ & [B, H, N+1, L/H] & 0.87\\
   Softmax & $\mQ \mK^T$ & [B, H, N+1, N+1] & 0.89\\
   GELU & & [B, N+1, L$\times D_{ffn}$] & 1.15\\		
        \bottomrule
    	\end{tabular}
    \caption{Memory cost for DeiT-S ($patch\_size=16, N=196, L=384, H=6, D_{ffn}=4$) model with a batch size B=1 and input image dimension of $224\times224$}
    \vspace{-6mm}
        \label{tab:memory_cost}
  \end{center}
\end{table}
From Table \ref{tab:memory_cost}, it is clear that activation memory is largely dependent on the number of input tokens. Softmax input exhibits quadratic dependence, while all saved activations exhibit linear dependence. With this motivation, we employ an efficient token dropping mechanism, where we selectively drop uninformative tokens, achieving over linear reduction in activation memory without sacrificing accuracy.

\section{Proposed Approach}
\subsection{Block Reprogramming in ViTs}
To improve the SoTA in fine-tuning ViTs, we propose Block Selective Reprogramming (BSR) which selectively trains a small fraction of blocks of a pre-trained model coupled with token dropping.
\begin{figure*}
    \centering
    \includegraphics[trim = 20 520 20 50, clip, width=0.84\textwidth]{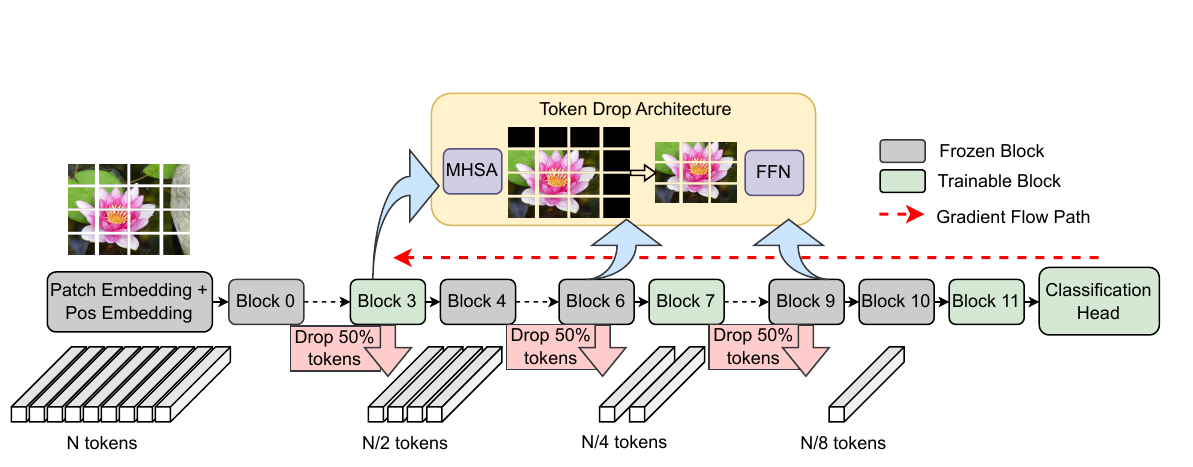}    
    \caption{Block selective reprogramming framework for a ViT model with 12 layers. The token drop locations are set at $4^{th}$, $7^{th}$, and $10^{th}$ blocks, where 50\% of incoming tokens are dropped based on token importance calculated by the MHSA module. The $4^{th}$, $8^{th}$, and $12^{th}$ blocks along with the classification head are trainable. The gradient propagation does not occur beyond the last trainable block.}
    \label{fig:block_diagram}
\end{figure*}
Token-dropping approaches have been widely studied in ViTs for latency and energy improvement \cite{liang2022not, bolya2022token}. However, token dropping has not been used in the context of activation memory reduction for on-device training. In this work, we couple token dropping with frozen-weight training, obtaining activation memory reduction up to $6\times$. Our token-dropping method is inspired by EViT \cite{liang2022evit}. We calculate token importance using self-attention scores and fuse low importance tokens based on a token drop rate. Thus, fewer tokens are passed on to the trainable blocks, significantly reducing activation memory. 

\begin{figure*}
  \centering
  \begin{subfigure}{0.27\linewidth}
    \includegraphics[width=\linewidth]{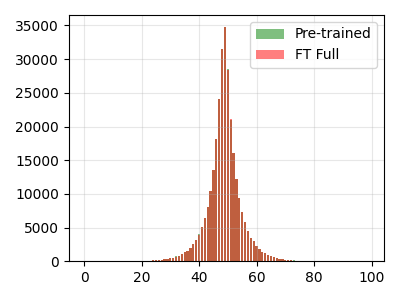}
    \caption{Patch embedding}
  \end{subfigure}
  \begin{subfigure}{0.27\linewidth}
    \includegraphics[width=\linewidth]{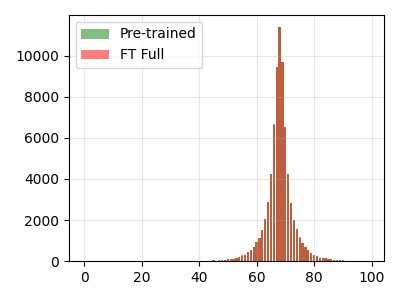}
    \caption{Positional embedding}
  \end{subfigure}
   \begin{subfigure}{0.27\linewidth}
    \includegraphics[width=\linewidth]{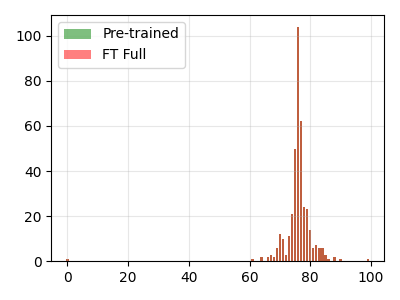}
    \caption{Classification token}
  \end{subfigure}
    \vspace{-2mm}
  \caption{Distribution of patch embedding, positional embedding and classification token of two DeiT-S: a pre-trained ImageNet model and a fully fine-tuned model on CIFAR-10}
  \label{fig:hist_plot}
\end{figure*}

The token importance is calculated by the MHSA module of a block
using the attention from the classification token to all other tokens, given by a score $\mS_{token}$ as follows
\begin{equation}
    {\mS_{token}} = \frac{\vq_{class} \mK^T}{\sqrt{d}}.
    \label{eq:stoken}
\end{equation}
The first row of the attention map $\mQ \mK^T$ is the dot product between the query vector obtained from classification token ($\vq_{class}$) and key matrix. We use this dot product to calculate token importance because tokens in the value matrix are linearly combined according to these scores ($\mS_{token}$) to predict the output class. The top-K tokens according to the importance score $\mS_{token}$ are preserved, $K$ being determined by the token drop rate, while the unimportant ones are fused and passed onto the subsequent FFN layer. The tokens are incrementally discarded across the length of the network through blocks termed token drop locations.
Since the classification token shows an exactly identical distribution for a pre-trained and fully fine-tuned model \ref{fig:hist_plot}, we leverage the pre-trained classification token for importance calculation. This alleviates the need to transmit gradients all the way back to the start of the network, thereby saving memory.

\subsection{Understanding the Gradient Flow}
In this section, we derive the gradient flow for residual learning and BSR, and justify why we choose to fine-tune parts of the main network instead of inserting residual blocks. We further refute Rep-Net's \cite{yang2022rep} claim that activation memory reduction is obtained because of an additive relation between trainable and frozen blocks.
\label{sec:grad_flow}

\begin{figure}[htbp]
    \centering
    \includegraphics[trim = 50 350 50 150, clip, width=0.45\textwidth]{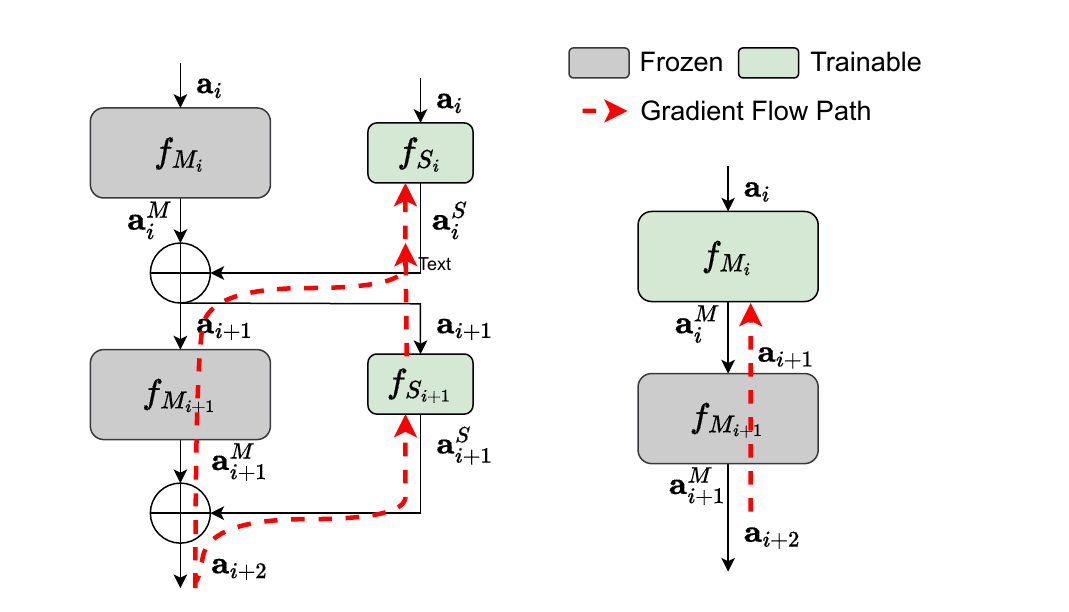}    
    \caption{\textit{Left:} Gradient flow in a residual learning architecture like \cite{yang2022rep} \textit{Right:} Gradient flow in our block selective reprogramming approach}
    \label{fig:residual_learning}
\end{figure}

In Figure  \ref{fig:residual_learning}, on the left, we consider a residual learning network like Rep-net, where the frozen $i^{th}$ main block, modeled as $f_{M_{i}}$ exchanges features with a trainable side block $f_{S_{i}}$. The input and output activations of the $i^{th}$ blocks are denoted by $\act_{i}$ and $\act_{i+1}$. The individual outputs of the main and side blocks are denoted by 
$\act_{M_{i}}$ and $\act_{S_{i}}$, where $\act_{M_{i}} = f_{M_{i}}(\act_{i})$ and $\act_{S_{i}} = f_{S_{i}}(\act_{i})$. The derivative of block $f_k$ is given by $f'_k$. 
The derivative of the loss $\loss$ with respect to the trainable weights $\weight_{S_{i}}$ of the $i^{th}$ side block $S_{i}$ is given as follows,
\begin{align}
    \nonumber
    \frac{\partial \loss}{\partial \weight_{S_{i}}} &= \frac{\partial \loss}{\partial \act_{i+2}} \frac{\partial \act_{i+2}}{\partial \weight_{S_{i}}} \\
    \nonumber
    =\frac{\partial \loss}{\partial \act_{i+2}} &\frac{\partial }{\partial \weight_{S_{i}}} \left(\act_{i+1}^{M}+\act_{i+1}^{S}\right) \\
    \nonumber
    =\frac{\partial \loss}{\partial \act_{i+2}} &\frac{\partial }{\partial \weight_{S_{i}}} \left(f_{M_{i+1}}\left(\act_{i+1}\right)+f_{S_{i+1}}\left(\act_{i+1}\right)\right) \\
    \nonumber
    = \frac{\partial \loss}{\partial \act_{i+2}} &\frac{\partial }{\partial \weight_{S_{i}}} \left(f_{M_{i+1}}\left(\act_{i}^{M}+\act_{i}^{S}\right)+f_{S_{i+1}}\left(\act_{i}^{M}+\act_{i}^{S}\right)\right) \\
    \label{eqn:grad_prop_res}
    =\frac{\partial \loss}{\partial \act_{i+2}}&\frac{\partial \act_{i}^{S}}{\partial \weight_{S_{i}}}\left({\color{mydarkred}f'_{M_{i+1}}(\act_{i}^{M}+\act_{i}^{S})}+f'_{S_{i+1}}\left(\act_{i}^{M}+\act_{i}^{S}\right)\right)
\end{align}

where the last step of the derivation utilizes the fact that $\act^{M}_{i}$ is independent of $\weight_{S_{i}}$.

Equation \ref{eqn:grad_prop_res} clearly shows that the gradient of weights of the trainable side block depends on the gradient through the subsequent main block ${\color{mydarkred}f'_{M_{i+1}}(\act_{i+1})}$.  Because the derivative of the loss depends on the input activation for the non-linear components, we still need to store these input activations along the main path. 

The advantage of freezing weights in CNNs stems from the fact that the derivative of the loss for linear layers like convolution need only weights, and not their input activation, as given by Equation \ref{eq:linear_backward}. ReLU, although non-linear, only needs to store a binary mask. Therefore, unlike the claims made by Rep-Net, the additive relationship between frozen and trainable blocks does not play a role in activation memory reduction. The memory reduction is obtained by reducing the activation dimension at the input of trainable blocks feeding downsampled versions of the activation to residual blocks. 
This advantage, however, is lost for ViTs where each encoder block consists of self-attention, softmax and GELU, each of which requires storing the activation input for gradient propagation (Equation \ref{eq:non_linear_backward}). Therefore, introducing residual blocks further increases the activation memory, as demonstrated through our experiments detailed in Section \ref{sec:residual}. Moreover, in residual learning the forward pass occurs through both the frozen main blocks and the trainable residual blocks, resulting in an increase in parameters and compute costs, both during training and inference. 

In Figure \ref{fig:residual_learning}, on the right, we show the gradient flow in BSR, where the $i^{th}$ main block is made trainable. The derivative of the loss $\loss$ with respect to the trainable weights $\weight_{M_{i}}$ of main block $M_{i}$ is given as follows
\begin{align}
\label{eqn:grad_prop_ours}
    \frac{\partial \loss}{\partial \weight_{M_{i}}} =& \frac{\partial \loss}{\partial \act_{i+2}} \frac{\partial \act_{i+2}}{\partial \weight_{M_{i}}} = \frac{\partial \loss}{\partial \act_{i+2}}\frac{\partial \act_{i}^{M}}{\partial \weight_{M_{i}}}\left(f'_{M_{i+1}}(\act_{i}^{M})\right)
\end{align}
The lack of a residual path simplifies Equation \ref{eqn:grad_prop_ours} and implies only activations along the main path need to be stored.

\subsection{Design Choice Discussion}
\noindent\textbf{Discussion 1} \textit{Which blocks are more essential during fine-tuning for transferring to small-scale datasets?}

We perform an analysis of intermediate features between an ImageNet pre-trained model and a fully fine-tuned model on CIFAR-10 dataset to observe which layers show the maximum differences. The layer statistics for the three trainable elements before the encoder blocks, patch embedding, positional embedding, and classification token, have been presented in Figure \ref{fig:hist_plot}. Surprisingly, we observe that these elements have an exactly identical distribution for the pre-trained and fine-tuned models. This is not obvious, particularly for the classification token, which is responsible for aggregating information for final classification. In general, fine-tuning the last layers of a model is more important \cite{chatfield2014return, donahue2014decaf, sharif2014cnn}, which is why in transfer learning applications, only the last layers are often fine-tuned. For CNNs, the first few layers or feature extraction layers are mostly similar across tasks. This brings us to our next discussion, namely, is only fine-tuning the last few layers sufficient in terms of accuracy and activation memory? 

\vpar{Discussion 2} \textit{Is fine-tuning the last few blocks enough?}

Table \ref{tab:last_blocks} presents results for training the last two, three, and four blocks, with and without token pruning. We observe that training blocks only towards the end cannot close the accuracy gap, even without token pruning. With token pruning, because only a few tokens are left towards the end, we observe a significant drop in accuracy. This necessitates an intelligent selection of trainable blocks and token drop locations, balancing between accuracy and activation memory to achieve an optimal trade-off.
\begin{table}[htbp]
\small\addtolength{\tabcolsep}{-0pt}
    \centering
    \begin{tabular}{ccccc}
    \toprule
        \textbf{Trainable} & \textbf{Token}  & \textbf{Train}  & \textbf{Reduce}  & \textbf{Accuracy} \\
        \textbf{Blocks} & \textbf{Dropping} & \textbf{Memory} & \textbf{Ratio} & \\
        \midrule
        Baseline & & 8649 & 1$\times$ &98.48\\
        \midrule
        $\left(10, 11\right)$ & \xmark & 1477 & 5.8$\times$ & 96.14 \\ 
        $\left(10, 11\right)$ & \cmark & 215 & 40.2$\times$ & 93.93 \\
        \midrule
        $\left(9, 10, 11\right)$ & \xmark & 2187 & 3.9$\times$ & 96.35 \\                
        $\left(9, 10, 11\right)$ & \cmark & 333 & 25.9$\times$ & 94.81   \\
        \midrule
        $\left(8, 9, 10, 11\right)$ & \xmark & 2896 & 2.9$\times$ & 96.81\\                
        $\left(8, 9, 10, 11\right)$ & \cmark & 502 & 17.2$\times$ & 95.23\\
        \bottomrule
    \end{tabular}
    \caption{Training last few blocks of a DeiT-S model on CIFAR-10 Dataset with and without token pruning}
    \label{tab:last_blocks}
\end{table}

\vpar{Discussion 3} \textit{Relative Positioning of Token Drop Locations and Trainable Blocks}
\label{disc:relative_pos}
The overall token drop rate of the network depends on the token drop locations and the fraction of tokens dropped in those locations. For example, if a higher token drop rate is chosen, placing token drop locations towards the end of the network will have the same effect as placing token drop locations in the shallower layers with a lower drop rate (Table \ref{tab:ablation2}). EViT \cite{liang2022evit} shows that the token dropping approach causes significant performance degradation when tokens are dropped before the $3^{rd}$ layer, since transformer models are not able to identify important tokens that early. They further observe that after the $3^{rd}$ layer, irrespective of drop location and drop rate, the network exhibits a stable performance for the same level of overall token reduction. We verify this observation in Section \ref{sec:ablation}.
We place token drop locations at the $4^{th}$, $7^{th}$, and $10^{th}$ blocks with a token drop rate of 0.5, resulting in an overall token reduction of 50\%. With this token drop configuration, the position of trainable blocks is varied to find a suitable trade-off between accuracy and training memory. An ablation for trainable block positions is presented in Table \ref{tab:ablation1}. We make several interesting observations.
\begin{itemize}
    \item Placing trainable blocks before token drop locations improves accuracy, but heavily costs training memory.
    \item Placing all trainable blocks after token drop locations yields significant memory advantages but is accompanied with a drop in accuracy.
    \item Training blocks at uniformly distributed depths performs better than solely training the last blocks.
\end{itemize}
Based on these observations, we keep 3 out of 12, i.e., the $4^{th}$, $8^{th}$ and $12^{th}$, blocks as trainable. The $4^{th}$ block is also a token drop location, where all $N$ tokens are processed by the MHSA for token importance calculation, passing on $N/2$ tokens to the FFN layer. The $8^{th}$ and the $12^{th}$ blocks process only $N/4$ and $N/8$ tokens respectively.

\section{Experimental Results}
\textbf{Models and Datasets:}
We demonstrate results on DeiT-S \cite{deit} and ViT-B \cite{dosovitskiy2020image}. Fine-tuning is performed on models pre-trained on ImageNet1k for DeiT-S, and ImageNet22k for ViT-B. MoE models are constructed from an Imagenet pre-trained ViT-B backbone, with three MoE layers, each consisting of 16 experts. Transfer learning performance is shown on five datasets: CIFAR-10, CIFAR-100 \cite{cifar}, Flowers \cite{flowers102}, Pets \cite{pets}, and Food \cite{food101}. The estimated memory results are calculated following \cite{cai2020tinytl, jiang2022back} and the on-device memory is measured on NVIDIA RTX A6000 GPUs.

\vpar{Training Hyperparameters:}
ViT-B models are trained in Pytorch following the training settings in \cite{jiang2022back}. ViT-B models are trained using SGD optimizer for 20k steps with cosine learning rate decay and initial learning rate tuned for each dataset. ViT MoE models are trained using SGD for 8k steps with an initial learning rate of 0.01. DeiT-S models are trained using AdamW optimizer for 50 epochs using cosine learning rate decay. The default image size is set to 224 and default patch size is set as 16.  

\subsection{Results and Analysis}
ViT-B and DeiT-S models for our approach are configured according to the settings described in Discussion 3 (Section \ref{disc:relative_pos}) with an overall token drop rate of 50\%.
Table \ref{tab:deits}  presents results for DeiT-S on CIFAR-10 and CIFAR-100. Our approach achieves $\textbf{6.03}\times$ training memory reduction, while showing accuracy degradation of $0.9\%$ on CIFAR-10 and $3.23\%$ on CIFAR-100. We also achieve a FLOPS reduction of ${\sim2}\times$. 
\begin{table}[htbp]
\small\addtolength{\tabcolsep}{-2pt}
    \centering
    \begin{tabular}{c|cc|c|cc}
    \toprule
    \textbf{Method} & \textbf{Train}         & \textbf{Reduce}  & \textbf{FLOPS}   & \textbf{CIFAR-10} & \textbf{CIFAR-100}\\
                    & \textbf{Mem.}   & \textbf{Ratio}   & \textbf{(GMacs)} &                   & \\
    \midrule
    FT-Full           & 8649 MB&  1$\times$ & 589 & 98.48 & 88.79\\
    FT-Last           & 369 MB  & 23.4$\times$ & 589 & 89.9 & 75.3\\
    Ours              & 1433 MB & 6.03$\times$ & 295 & 97.5 & 85.56\\
    \bottomrule
    \end{tabular}
    \caption{BSR for DeiT-S on CIFAR-10 and CIFAR-100 with batch size 128}
    \label{tab:deits}
\end{table}

\begin{table*}[ht!]
\small\addtolength{\tabcolsep}{-2pt}
    \centering
    \begin{tabular}{c|cc|cc|cc|ccccc}
    \toprule
    \textbf{Method} & \textbf{Train}         & \textbf{Reduce} & \textbf{On-device}         & \textbf{Reduce} & \textbf{FLOPS} $\downarrow$  & \textbf{Reduce} & \multicolumn{5}{c}{\textbf{Accuracy}($\%$)$\uparrow$} \\
                    & \textbf{Mem. (MB)} $\downarrow$  & \textbf{Ratio} $\uparrow$  & \textbf{Mem. (MB)} $\downarrow$  & \textbf{Ratio} $\uparrow$  & \textbf{(GMacs)} & \textbf{Ratio} $\uparrow$ & \textbf{CIFAR-10} & \textbf{CIFAR-100} & \textbf{Flowers} & \textbf{Pets} & \textbf{Food} \\
    \midrule
    FT-Full           & 17388 &  1$\times$  & 19098 & 1$\times$     & 2249 & 1$\times$ & 98.98 & 93.1 & 99.2 & 93.7 & 90.5\\
    FT-Last           & 525   & 33$\times$  & 1629  & 11.7$\times$  & 2249 & 1$\times$ & 96.1  & 84.5 &  99.02 & 92.2 & 84.1 \\
    \midrule
    BackRazor$@$80 \% \cite{jiang2022back} & 4565 & 4.2$\times$ & 8501 & 2.66$\times$ &2249 & 1$\times$ & \underline{98.9} & \underline{92.9}  & \underline{99.4} & \underline{93.8} & \underline{90.5} \\
    BackRazor$@$95\% \citep{jiang2022back} & 3496 & 5.5$\times$ & 7189 & 3.15$\times$ & 2249 & 1$\times$ &98.8 & 90.0 & 99.4 & 92.4 & 88.7 \\
    \midrule
    Ours              & \textbf{2938} & \textbf{5.92$\times$} & \textbf{4342} & \textbf{4.4$\times$} & \textbf{1129} & \textbf{2}$\times$  &98.3  & 91.02  & 99.1 & 92.4  & 88.4 \\
    \bottomrule
    \end{tabular}
    \caption{Comparison of BSR with other transfer learning approaches for ViT-B with batch size 128. The values for Back Razor are taken directly from the paper \cite{jiang2022back}. The training memory reduction ratios for Back Razor are calculated with respect to their reported baseline. The best accuracy other than FT-Full is underlined. Back Razor$@$80\% gives the best accuracy with limited memory reduction. BSR yields the highest reduction in estimated and measured memory and FLOPS (highlighted in bold).}
    \label{tab:sota_comparison}
\end{table*}

In Table \ref{tab:sota_comparison}, we compare full fine-tuning (FT-Full), fine-tuning only the last layer or the classification head (FT-Last), Back Razor \cite{jiang2022back}, and our approach. The training memory, on-device memory, and FLOPS are calculated for CIFAR-100. We compare with Back Razor for 80\% and 95\% activation sparsity. FT-Last provides the highest memory benefits, while performing reasonably well on datasets like Flowers, but suffers substantial degradation for CIFAR-100 and Food. Some of our FT-Full accuracies are marginally lower than baseline accuracies reported in Back Razor. For example, for Flowers-102, our FT-Full accuracy is 99.2\% as opposed to their reported accuracy of 99.5\%. Both Back Razor and our method suffer 0.1\% degradation in Flowers-102. For Pets, the accuracy of our method is at par with Back Razor$@$95\%, although our FT-Full accuracy is lower by 0.4\%, implying our method suffers 0.4\% less degradation. Our approach outperforms Back Razor$@$95\% for CIFAR-100 and Pets by 1.02\% and 0.4\%, and performs at par on Flowers dataset. We provide greater training memory reduction than Back Razor$@$95\% while simultaneously achieving $2\times$ FLOPS reduction.  

\begin{figure}[htb]
\begin{minipage}[b]{.45\linewidth}
  \centering
  \centerline{\includegraphics[width=4.3cm]{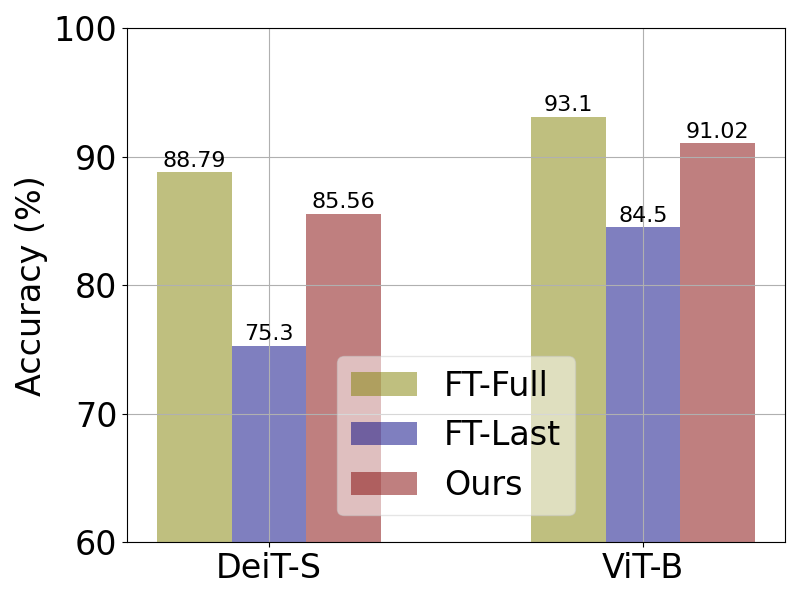}}
\end{minipage}
\hfill
\begin{minipage}[b]{0.48\linewidth}
  \centering
  \centerline{\includegraphics[width=4.3cm]{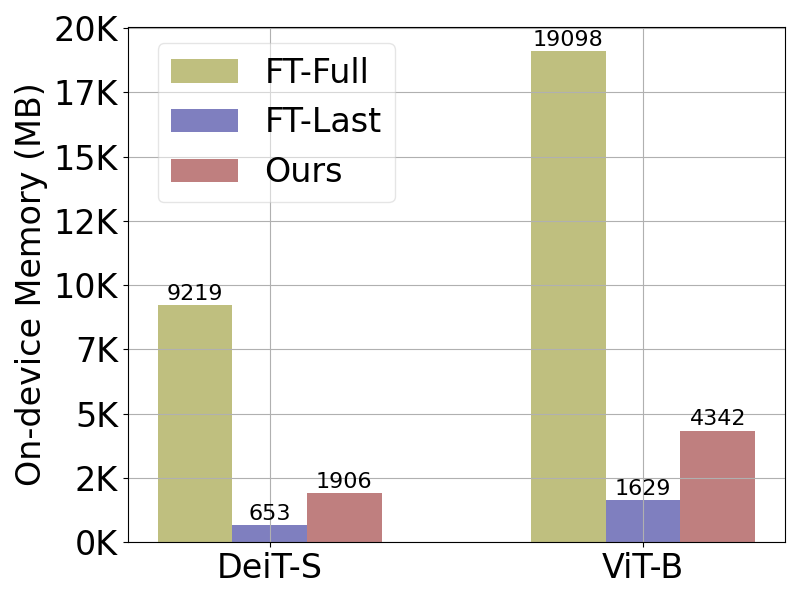}}
\end{minipage}
\vspace{-0.2cm}
\caption{Accuracy and on-device memory for DeiT-S and ViT-B models on CIFAR-100}
\label{fig:ondevice}
\end{figure}

\vpar{On-device Memory:}
The accuracy and on-device memory for FT-Full, FT-Last, and BSR on CIFAR-100 are presented in Figure \ref{fig:ondevice}. FT-Last provides the highest memory reduction but suffers substantial ($\sim$10\% on DeiT-S) accuracy degradation. Our approach provides on-device memory reduction of $4.8\times$ for DeiT-S model and $4.4\times$ for ViT-B and outperforms Back Razor by $1.4\times$.  


\vpar{Residual Learning for CNNs and ViTs}
\label{sec:residual}
We also investigate residual learning inspired from Rep-Net for ViTs. Three trainable side blocks are introduced while the pre-trained backbone is kept frozen. The output activations of main and side blocks are added together, similar to Figure \ref{fig:residual_learning}. BSR outperforms residual learning in accuracy, training memory and FLOPS, as shown in Table \ref{tab:residual}. 

For testing BSR on CNNs, we train only the last blocks of layers 2, 3, 4 in ResNet-50. We outperform FT-Full and achieve similar performance as Rep-Net with lesser FLOPS and training memory. 
\begin{table}[htbp]
\small\addtolength{\tabcolsep}{-1pt}
    \centering
    \begin{tabular}{c|c|c|c|c}
    \toprule
    \textbf{Model} & \textbf{Method} & \textbf{Accuracy} $\uparrow$ & \textbf{Activation} $\downarrow$  & \textbf{FLOPS} $\downarrow$ \\
          &        &    (\%)      & \textbf{Mem.(MB)} & \textbf{(GMacs)}\\
    \midrule
    \multirow{3}{*}{ResNet-50} & FT-Full & 95.13 & 88.4 & 4.1\\
     & Rep-Net & 96.37 & 9.91 & 6.2\\
     & Ours & 96.21 & 8.02 & 4.1\\
    \midrule
    \multirow{3}{*}{DeiT-S}  & FT-Full &98.48 & 66.9 & 4.6 \\
     & Rep-Net  & 97.81 & 40.1 & 5.7 \\ 
     & Ours & 98.16 & 34.3 & 4.6 \\
    \bottomrule
    \end{tabular}
    \caption{Residual learning and BSR for CNNs and ViTs on CIFAR-10. The activation memory and FLOPS are reported for a batch size of 1.}
    \label{tab:residual}
\end{table}

\begin{table*}[htbp]
\small\addtolength{\tabcolsep}{-0pt}
    \centering
    \begin{tabular}{c|c|cc|cc|ccccc}
    \toprule
    \textbf{Method} & \textbf{Model} & \textbf{On-device}         & \textbf{Reduce} & \textbf{FLOPS} $\downarrow$  & \textbf{Reduce} & \multicolumn{5}{c}{\textbf{Accuracy}($\%$)$\uparrow$} \\
                       & & \textbf{Mem. (MB)} $\downarrow$  & \textbf{Ratio} $\uparrow$  & \textbf{(GMacs)} & \textbf{Ratio} $\uparrow$ & \textbf{CIFAR-10} & \textbf{CIFAR-100} & \textbf{Flowers} & \textbf{Pets} & \textbf{Food} \\
    \midrule
    STL & Baseline  & 19098 & 1$\times$     & 2249 & 1$\times$ & 98.98 & 93.1 & 99.2 & 93.7 & 90.5\\
    STL & Ours & \textbf{4342} & \textbf{4.4$\times$} & \textbf{1129} & \textbf{2}$\times$  &98.3  & 91.02  & 99.1 & 92.4  & 88.4 \\
    \midrule
    MoE & Baseline & 30722 & 1$\times$ & 1801 & 1$\times$  & 98.93 & 92.77 &	98.33 & 92.21 & 90.12  \\
    MoE & Ours & \textbf{13244} & \textbf{2.32$\times$} & \textbf{1237} & \textbf{1.5}$\times$  &98.72  & 91.99  & 98.24 & 93.24  & 89.34 \\
    \bottomrule
    \end{tabular}
    \caption{BSR for MoE models with ViT-B backbone with batch size 128}
    \label{tab:moe}
\end{table*}

\subsection{Extending BSR for Mixture-of-Experts}
Multi-task learning (MTL) \cite{liang2022mvit, chen2023adamv} is becoming pivotal for edge intelligence because of its ability to adapt to tasks dynamically with minimum overhead. MTL models learn a shared representation for different tasks, thereby avoiding the overhead of training as well as storing separate models. Mixture-of-experts (MoE) \cite{chen2023sparse, liang2022mvit} constitute a new paradigm in MTL that separates the parameter space by activating parts of the model based on task and input tokens, providing improved MTL performance. In this section, we demonstrate that our approach can be easily extended for fine-tuning MoE models. To the best of our knowledge on-device training has not been previously explored for MoE models.

For ViT MoE, the FFN in ViT is replaced by a MoE layer. A MoE layer consists of several experts represented as MLPs. A task-dependent router sparsely activates a subset of experts for each input token. We use a MoE model with a ViT-B backbone. We only replace the MLP layers for the three blocks (corresponding to the location of trainaable blocks in BSR) with MoE layers. The MoE layers have 16 expert candidates, out of which top four candidates for each token are selected by the router, adding up the results to generate the output of the MoE layer. The results for BSR fine-tuning of MoE are presented in Table \ref{tab:moe}. Single task learning (STL) consists of separate models for each task. We observe that BSR performs remarkably well for MoE fine-tuning causing accuracy degradation of only 0.21\%, 0.78\%, 0.09\% and 0.78\% over the MoE baseline for CIFAR-10, CIFAR-100, Flowers and Food while outperforming the baseline by 1.04\% for Pets. We obtain on-device memory reduction by $2.32\times$ and FLOPs reduction by $1.5\times$. Notably, our MoE models significantly outperform the STL models in most cases. 

\subsection{Ablation Studies}
\label{sec:ablation}
\noindent\textbf{Number and Position of Trainable Blocks}
Table \ref{tab:ablation1} presents an ablation on the number and position of trainable blocks. The indices runs from 0 to 11 corresponding to the model depth of twelve layers. Training with four blocks provides no significant advantages over three blocks. In fact, three blocks often performs better than four blocks, as seen for training blocks [4, 7, 11] vs [4, 7, 10, 11] or [2, 5, 8, 11] vs [2, 5, 8]. Training with two blocks causes much higher degradation. Therefore we choose the number of trainable blocks as three. The position of trainable blocks with respect to the token drop locations provides an accuracy memory trade-off, as discussed in Section \ref{disc:relative_pos}. 
\begin{table}[htbp]
\small\addtolength{\tabcolsep}{-1pt}
    \centering
    \begin{tabular}{c|c|cc|c}
    \toprule
         \textbf{\#Trainable} & \textbf{Indices} & \textbf{Train}       & \textbf{Reduce} & \textbf{Accuracy}\\
         \textbf{Blocks}      &         & \textbf{Mem. (MB)} & \textbf{Ratio}  & $(\%)$ \\
         \midrule
        \multirow{4}{*}{4}  & $\left[3, 6, 9, 11\right]$ & 1526 & 5.7$\times$ & 97.44\\  
                            & $\left[2, 5, 8, 11\right]$ & 2809 & 3.1$\times$ & 97.82\\
                            & $\left[4, 7, 10, 11\right]$ & 1132 & 7.6$\times$ & 97.17\\
                            & $\left[8, 9, 10, 11\right]$ & 502 & 17.2$\times$ & 95.23\\
                            & $\left[0, 2, 5, 8\right]$ & 3705 & 2.3$\times$ & 97.75\\
        \midrule
        \multirow{5}{*}{3}  & $\left[3, 7, 11\right]$ & 1433 & 6.03$\times$ & 97.43\\
                            & $\left[4, 7, 11\right]$ & 1079 & 8.0$\times$ & 97.34\\
                            & $\left[2, 7, 11\right]$ & 2731 & 3.2$\times$ & 97.81\\
                            & $\left[9, 10, 11\right]$ & 333 & 25.9$\times$ & 94.81\\
                            & $\left[2, 5, 8\right]$ & 2785 & 3.1$\times$ & 97.86\\
        \midrule
        \multirow{3}{*}{2}  & $\left[4, 11\right]$ & 986 & 8.8$\times$ & 96.80\\
                            & $\left[7, 11\right]$ & 460 & 18.8$\times$ & 95.84\\
                            & $\left[10, 11\right]$ & 215 & 40.2$\times$ & 93.93\\
    \bottomrule
    \end{tabular}
    \caption{Ablation study on the number and position of trainable blocks for DeiT-S on CIFAR-10. The token drop locations are kept frozen at the $4^{th}$, $7^{th}$, and $11^{th}$ blocks.}
    \label{tab:ablation1}
\end{table}

\vpar{Token Drop Rate and Drop Locations:}
Table \ref{tab:ablation2} presents an ablation study on token drop rates and locations such that the overall drop rate is around 50\%. We observe that for similar drops there is minor accuracy variation (${\sim}0.1\%$) while the memory and FLOPs reduction are also similar. 
\begin{table}[htbp]
\small\addtolength{\tabcolsep}{-0.5pt}
    \centering
    \begin{tabular}{c|c|cc|c}
    \toprule
    \textbf{Drop} & \textbf{Drop} & \textbf{Train}       & \textbf{FLOPs} & \textbf{Accuracy}\\
    \textbf{Rate} & \textbf{Indices} & \textbf{Mem. (MB)} & \textbf{(GMACs)} & $(\%)$ \\
    \midrule
    0.3 & [1,3,5,7,9] & 1363 & 270 & 97.54\\
    0.5 & [3,6,9] & 1433 & 295 & 97.42\\
    0.7 & [5,7,9] & 1673 & 311 &97.43\\
    \bottomrule
    \end{tabular}
    \caption{Ablation study on the token drop rate and locations for DeiT-S on CIFAR-10. The trainable block positions are fixed at the $4^{th}$, $8^{th}$, and $12^{th}$ blocks.}
    \label{tab:ablation2}
\end{table}

\section{Conclusions}
We propose BSR, an approach for efficient on-device training of ViTs that couples selectively fine-tuning a small fraction of blocks of pre-trained models on downstream tasks with token dropping based on self-attention scores of primarily frozen weights. Unlike existing on-device learning approaches that reduce training memory at the cost of increased computational overhead, BSR reduces compute cost by $2\times$ while reducing training time by ${\sim}2.5\times$ and memory up to $6.03\times$. We outperform SoTA approaches in on-device training memory reduction by 1.4$\times$ while maintaining similar performance. We further demonstrate the effectiveness of the approach for MoE models in MTL applications. Our approach is orthogonal to activation quantization which can be used to obtain further reductions in activation memory.

{
    \small
    \bibliographystyle{ieeenat_fullname}
    \bibliography{main}
}


\end{document}